\documentclass{article}
\usepackage{iclr2027_conference,times}

\usepackage{graphicx}
\usepackage{xcolor}
\usepackage{amsmath,amssymb,amsfonts}
\usepackage{mathtools}
\usepackage{bm}
\usepackage{bbm}
\usepackage{booktabs}
\usepackage{float}
\usepackage{adjustbox}
\usepackage{placeins}
\usepackage{colortbl}
\usepackage{xspace}
\usepackage{array}
\usepackage{multirow}
\usepackage{makecell}
\usepackage{tabularx}
\usepackage{longtable}
\usepackage{pifont}
\usepackage{tikz}
\usetikzlibrary{arrows.meta,calc,decorations.text,fit,positioning,shapes.geometric}
\usepackage{hyperref}
\usepackage{xurl}
\usepackage{threeparttable}

\providecommand{\ecyes}{\textcolor{pathgreen}{\ding{51}}}
\providecommand{\ecno}{\textcolor{pathred}{\ding{55}}}
\providecommand{\ecpartial}{\textcolor{pathgold}{$\triangle$}}

\newcolumntype{L}{>{\raggedright\arraybackslash}X}
\newcolumntype{C}{>{\centering\arraybackslash}p{0.057\textwidth}}

\definecolor{pathblue}{HTML}{176B87}
\definecolor{pathgreen}{HTML}{3B7A57}
\definecolor{pathred}{HTML}{B6413A}
\definecolor{pathgold}{HTML}{B57918}
\definecolor{pathlight}{HTML}{EDF4F6}
\definecolor{pathhema}{HTML}{5A4778}
\definecolor{pathhemawash}{HTML}{EEEAF4}
\definecolor{patheosinwash}{HTML}{F8E7ED}

\newcolumntype{Y}{>{\raggedright\arraybackslash}X}
\newcolumntype{P}[1]{>{\raggedright\arraybackslash}p{#1}}

\newcommand{\figref}[1]{Fig.~\ref{#1}}

\DeclareRobustCommand{\ours}{\textsc{SciRigor}\xspace}
\newcommand{\papertitle}{\ours: Evaluating Open-Ended Scientific Analysis Beyond Final Scores}
\title{\papertitle}

\author{
Bowen Liu\textsuperscript{1,*}\quad
Shuo Nie\textsuperscript{2,*}\quad
Bodong Du\textsuperscript{1,*}\quad
Xiaomeng Li\textsuperscript{1,\textdagger}\\
\textnormal{\small \textsuperscript{1}HKUST\quad
\textsuperscript{2}Alibaba Group}\\
\textnormal{\small \textsuperscript{*}Equal contribution.\quad
\textsuperscript{\textdagger}Corresponding author.}
}

\iclrfinalcopy

\hypersetup{
  pdftitle={\ours: Evaluating Open-Ended Scientific Analysis Beyond Final Scores},
  pdfauthor={Bowen Liu, Shuo Nie, Bodong Du, Xiaomeng Li}
}

\begin{document}

\maketitle
\fancyhf{}
\fancyhead[L]{\scriptsize\itshape \papertitle}
\fancyfoot[C]{\thepage}
\renewcommand{\headrulewidth}{0.4pt}

\begin{abstract}
Scientific coding agents increasingly perform long-horizon, open-ended
workflows that produce interdependent code, numerical results, figures, and
claims. Existing evaluations typically score workflow completion or final
artifacts in isolation. These scores do not establish whether a reported claim
is supported by the data, computation, and visualization produced in the same
run. We formulate \emph{evidence-grounded scientific analysis with multimodal
artifacts} as an evaluation task.
Given a scientific question, analysis-ready data, and study context, an agent
must produce an executable analysis, structured results, a visualization, and
atomic scientific claims. Success requires the required findings to be reported
and every primary claim to retain a verified same-run support path.
\textbf{\ours} couples a fine-grained evaluation framework with an
evidence-complete benchmark for this task. The framework
reconstructs generated artifacts as a typed evidence graph, evaluates artifact
fidelity separately from edge validity, scores claims over complete
computational and visual paths, and localizes the earliest unsupported relation.
Source-grounded alternative paths accommodate scientifically equivalent methods
and visual encodings, while weakest-link scoring exposes upstream errors behind
plausible downstream outputs. The benchmark comprises 100 cases drawn from
scientific articles published in journals such as \emph{Nature}, spanning six
domains and 17 subfields. Each case aligns the question, complete inputs,
executable source analysis, numerical results, source figure, atomic findings,
and provenance. We evaluate 11 agent/model configurations on \ours.
On full-benchmark runs, agents'
claims agree with their own results at nearly the same rate whether those results
are faithful to the scientific target or not (91.8\% versus 91.0\%). Yet no
system exceeds 62.6\% on the soft evidence-chain score or 18.0\% strict
whole-chain success. Internal coherence is therefore not sufficient evidence of
scientific correctness: support must be verified along the full data-to-claim
path.
\end{abstract}

\section{Introduction}
\label{sec:introduction}

Scientific coding agents are moving beyond isolated code generation toward
open-ended analysis workflows that produce numerical results, figures, and
scientific conclusions~\citep{chen2025scienceagentbenchrigorousassessmentlanguage,wang2026naturebenchcodingagentsmatch,wang2026firebenchevaluatingaiagents}.
Because these outputs may enter papers, reports, and downstream decisions, their
evidential validity matters as much as their surface plausibility.
These artifacts may look credible on their own while failing to support one
another. Consider a case in our collection that asks how food-delivery workload
changes under extreme heat (Appendix~\ref{app:case-food-delivery}). The source
regression stores an estimate and its confidence limits on a log scale, so a
correct analysis must apply $100[\exp(x)-1]$ to all three values. If an agent
transforms only the estimate, it can still state the correct-looking finding of
an 18.0\% increase (95\% CI 12.6--23.6\%), even though its structured interval
and plotted uncertainty band remain on the wrong scale. The sentence matches the
scientific target, but the submitted computation and figure do not support it.
Evaluating such workflows therefore requires more than final-output scoring,
which judges whether a returned artifact looks plausible or resembles a
reference without verifying how it was produced.

Existing benchmarks score complementary parts of this workflow. Scientific
programming benchmarks assess code, execution, and analysis
decisions~\citep{chen2025scienceagentbenchrigorousassessmentlanguage,zhang2025datascibenchllmagentbenchmark,gu2025bladebenchmarkinglanguagemodel};
research-agent benchmarks extend evaluation to hypotheses, experiments, and
multi-artifact outputs~\citep{majumder2024discoverybenchdatadrivendiscoverylarge,wang2026firebenchevaluatingaiagents,xu2026researchclawbenchbenchmarkendtoendautonomous};
and visualization or claim-auditing benchmarks examine figures and scientific
statements~\citep{ai2026scivisagentbenchbenchmarkevaluatingscientific,deng2026scifigqualbenchbenchmarkscientificfigure,yue2026factreviewevidencegroundedpeerreview}.
Across these settings, the scored object is typically a program, stage, artifact,
or final outcome. They therefore do not jointly establish whether a newly
generated claim is supported by the data, computation, and visualization
produced in the same run. This blind spot can accept a correct-sounding but
unsupported conclusion, while exact-reference matching can reject a defensible
alternative analysis.

We formulate \emph{evidence-grounded scientific analysis with multimodal
artifacts} as the post-experiment task between collected data and a reportable
conclusion. Given a scientific question, analysis-ready data, and study context,
an agent produces mutually supporting code, results, a figure, and atomic
claims. It may choose among defensible methods and visual encodings. Task success requires the
required findings to be reported and every primary claim to retain at least one
complete computational and visual support path. The path comprises observable
scientific artifacts and dependencies rather than
hidden reasoning or proof steps. We ask: \textbf{RQ1:} How faithfully do agents
construct each artifact? \textbf{RQ2:} Does agreement among an agent's own
artifacts imply agreement with the scientific target? \textbf{RQ3:} How often
does local success survive composition into a complete data-to-claim path?

A suitable evaluation must ground every scored artifact in executed data,
reconstruct relations among artifacts from runtime evidence, and accommodate
scientifically equivalent methods and visual encodings. \ours meets these
requirements with a typed evidence graph. During benchmark construction, source
code, data, results, figures, and findings define human-audited support paths.
During evaluation, the evaluator reconstructs the candidate graph from executed
artifacts, scores artifact fidelity separately from edge validity, assigns each
claim the strength of its best pre-specified support path under a weakest-link
rule, and localizes the earliest unsupported relation. Complete-path scoring
exposes plausible claims whose computational or visual support has already
failed, while source-grounded alternative paths admit valid non-reference
analyses.

We instantiate this framework in 100 evidence-complete cases drawn from
scientific articles published in journals such as \emph{Nature}, spanning six
domains and 17 subfields
(Figure~\ref{fig:benchmark-statistics}). Each case aligns a question with all
code-consumed data, executable source analysis, numerical results, a source
figure, atomic findings, and provenance. The agent receives only the question,
declared inputs, and study context; every admitted case passes two clean
reproductions and a final human audit. We evaluate 11 agent/model
configurations on \ours. On full-benchmark runs, agents'
claims agree with their own results at nearly identical rates whether those
results are faithful to the
scientific target or not (91.8\% versus 91.0\%), while the best strict
whole-chain success rate is only 18.0\%. These results show that internal
coherence is not sufficient evidence of scientific correctness.

Our contributions are:
\begin{itemize}
    \item \textbf{Task formulation.} We formulate evidence-grounded scientific
    analysis with multimodal artifacts. The task connects generated code,
    results, a figure, and atomic claims through verifiable same-run support
    (Section~\ref{sec:task-formulation}).
    \item \textbf{Evaluation framework.} We define the scored object as a
    verified data-to-claim support path, separate artifact fidelity from edge
    validity, and localize the earliest unsupported relation in a typed evidence
    graph
    (Section~\ref{sec:evaluation-framework}).
    \item \textbf{Benchmark.} We construct 100 evidence-complete
    cases spanning six domains and 17 subfields, with executable reconstruction,
    source-grounded alternative paths, and human-audited provenance
    (Section~\ref{sec:benchmark}).
    \item \textbf{Empirical diagnosis.} We evaluate 11 agent/model configurations
    and quantify where execution, scientific fidelity, self-consistency, and
    end-to-end evidence composition diverge
    (Section~\ref{sec:experiments}).
\end{itemize}

\section{Related Work}
\label{sec:related-work}

Scientific-agent benchmarks differ in their primary scored object: an
executable analysis, a completed research workflow, or an individual figure or
claim. \ours connects these levels by scoring a generated claim together with
the computation and visualization produced to support it.
Table~\ref{tab:benchmark-comparison} summarizes the most relevant comparisons.

\paragraph{Scientific programming and data-analysis benchmarks.}
Scientific programming benchmarks make executable code or analytical decisions
the primary unit. NatureBench~\citep{wang2026naturebenchcodingagentsmatch} tests
reproduction or improvement of published results in standardized environments,
while ScienceAgentBench~\citep{chen2025scienceagentbenchrigorousassessmentlanguage}
scores programs and their execution on paper-derived tasks.
BLADE~\citep{gu2025bladebenchmarkinglanguagemodel} evaluates open-ended choices
of variables, transformations, and statistical models, and
DataSciBench~\citep{zhang2025datascibenchllmagentbenchmark} evaluates practical
data-science agents through executable tasks and artifact-specific metrics.
These benchmarks verify whether an analysis is executable or scientifically
defensible, but do not extend the scored unit through a generated figure to its
reported claims.

\paragraph{Scientific discovery and end-to-end research agents.}
Research-agent benchmarks broaden the unit from one analysis to a multi-stage
workflow. DiscoveryBench~\citep{majumder2024discoverybenchdatadrivendiscoverylarge}
evaluates discoveries from supplied data;
FIRE-Bench~\citep{wang2026firebenchevaluatingaiagents} evaluates rediscovery from
research questions through experiments and conclusions; and
ResearchClawBench~\citep{xu2026researchclawbenchbenchmarkendtoendautonomous}
scores complete research artifacts with multimodal rubrics.
FrontierChallenge~\citep{su2026frontierchallengeevaluatingscientificworkflow} uses
deliverable-specific executable graders for fixed-input workflows. These
benchmarks evaluate whether a workflow delivers the required research outputs,
whereas \ours asks whether each generated claim retains a verified support
path within that completed workflow. SciDataBench~\citep{anonymous2026scidatabench}
further studies agents operating over scientific data APIs.

\paragraph{Visualization, claim verification, and evidence chains.}
Visualization and claim-verification benchmarks isolate individual links in the
same chain.
SciVisAgentBench~\citep{ai2026scivisagentbenchbenchmarkevaluatingscientific}
evaluates multi-step scientific analysis and visualization, while
SciFigQual-Bench~\citep{deng2026scifigqualbenchbenchmarkscientificfigure} scores
existing figures against captions and manuscript context.
FactReview~\citep{yue2026factreviewevidencegroundedpeerreview} audits claims
extracted from papers using literature and execution evidence. \ours instead
jointly evaluates newly generated results, figures, and claims, reconstructs
their dependencies from runtime evidence, and localizes the earliest unsupported
edge.

\begin{table*}[t]
\centering
\caption{Support coverage of scientific-agent benchmarks.}
\label{tab:benchmark-comparison}
\footnotesize
\setlength{\tabcolsep}{7.0pt}
\renewcommand{\arraystretch}{1.10}
\begin{adjustbox}{max width=\textwidth}
\begin{tabular}{l*{5}{c}}
\toprule
\textbf{Benchmark} & \textbf{Code} &
\textbf{Figure} &
\textbf{Generated claim} &
\textbf{Claim path} &
\textbf{First failure} \\
\midrule
ScienceAgentBench~\citep{chen2025scienceagentbenchrigorousassessmentlanguage}
  & \ecyes & \ecpartial & \ecno & \ecno & \ecno \\
BLADE~\citep{gu2025bladebenchmarkinglanguagemodel}
  & \ecyes & \ecno & \ecno & \ecpartial & \ecno \\
\addlinespace[2pt]
DiscoveryBench~\citep{majumder2024discoverybenchdatadrivendiscoverylarge}
  & \ecpartial & \ecno & \ecyes & \ecno & \ecno \\
DataSciBench~\citep{zhang2025datascibenchllmagentbenchmark}
  & \ecyes & \ecpartial & \ecpartial & \ecno & \ecno \\
FIRE-Bench~\citep{wang2026firebenchevaluatingaiagents}
  & \ecyes & \ecpartial & \ecyes & \ecno & \ecpartial \\
\addlinespace[2pt]
SciVisAgentBench~\citep{ai2026scivisagentbenchbenchmarkevaluatingscientific}
  & \ecyes & \ecyes & \ecno & \ecno & \ecno \\
\midrule
\rowcolor{pathlight}
\textbf{\ours} & \ecyes & \ecyes & \ecyes & \ecyes & \ecyes \\
\bottomrule
\end{tabular}
\end{adjustbox}
\vspace{2pt}
\parbox{\textwidth}{\raggedright
\textit{Symbols:} \ecyes{} = required and directly evaluated;
\ecpartial{} = subchain-, rubric-, contract-, or stage-level coverage;
\ecno{} = outside the primary scope.
\textit{Claim path} requires verification of same-run computational and visual
support; \textit{First failure} localizes the earliest unsupported evidence edge
within an instance.\par}
\end{table*}

\section{\ours}
\label{sec:scirigor}

\ours evaluates scientific analysis as a connected evidentiary object. It
defines what an agent must produce, reconstructs the dependencies among the
executed artifacts, and scores whether each reported claim retains complete
computational and visual support. This progression separates three questions
that final-output scores conflate: whether an artifact is faithful to the
scientific target, whether one artifact supports another, and whether those
local relations compose into an end-to-end data-to-claim path.

\subsection{Task Formulation}
\label{sec:task-formulation}

We formulate \emph{evidence-grounded scientific analysis with multimodal
artifacts} as the post-experiment task between collected data and a reportable
conclusion. Let
$q$ denote a scientific question, $\mathcal{D}$ the analysis-ready data, and
$\kappa$ the study context and measurement semantics. An agent $\mathcal{A}$
produces four interdependent scientific artifacts:
\begin{equation}
  \mathcal{A}(q,\mathcal{D},\kappa)
  \longmapsto
  \widehat{\mathcal{Y}}
  = (\widehat{T},\widehat{R},\widehat{V},\widehat{C}),
  \label{eq:task-formulation}
\end{equation}
where $\widehat{T}$ is an executable analysis, $\widehat{R}$ its structured
numerical results, $\widehat{V}$ a data-grounded visualization, and
$\widehat{C}$ a set of atomic scientific claims. Multimodality refers to the
generated visualizations and textual claims, whose support is traced through
code and numerical results. The joint output makes the
relations among artifacts part of the task rather than treating each artifact
as an independent endpoint.

The task is executable: running $\widehat{T}$ over $\mathcal{D}$ must regenerate
$\widehat{R}$, $\widehat{V}$, and $\widehat{C}$ and expose provenance
$\widehat{P}$ for evaluator verification. A successful submission covers the
required findings and gives every primary claim at least one verified
computational and visual support path through artifacts generated in the same
run. Pre-specified, source-grounded alternatives represent scientifically valid
choices of analysis method and visual encoding. Appendix~
\ref{app:protocol-tables}, Table~\ref{tab:five-file-contract}, maps these abstract
artifacts to the serialized submission contract.

\subsection{Evidence-Graph Evaluation}
\label{sec:evaluation-framework}

\ours constructs the reference support specification once, before model
evaluation. Curators execute the frozen source workflow and canonicalize its
consumed data views, transformations, statistical results, visual marks, and
atomic findings into semantic slots. Each required claim slot is linked to at
least one computational support set and one visual support set. Items within a
set are jointly required, whereas separate sets represent approved alternative
methods or visualizations. A final human audit verifies that the selected panel,
source passage, author code, consumed data, claim scope, and support sets address
the same scientific question. The resulting hidden specification determines the
relations to verify without prescribing candidate filenames, identifiers, code,
prose, or pixels.

At evaluation time, the submission becomes an executed dependency graph over
observable scientific artifacts rather than an agent-authored reasoning trace.
The evaluator reconstructs this graph; the agent does not submit it as an
additional artifact. Its core relations are
\begin{equation}
  D \rightarrow T \rightarrow R \rightarrow C,
  \qquad
  T \rightarrow V,\; R \rightarrow V,\; V \rightarrow C,
  \label{eq:evidence-graph}
\end{equation}
where $D$, $T$, $R$, $V$, and $C$ denote data assets, transformations,
statistical results, visual marks, and atomic claims. The evaluator instantiates
$D$ and $T$ from runtime traces, $R$ from regenerated candidate results and
independent numerical oracles, $V$ from captured plotting objects and arrays,
and $C$ from structured claim records. Here the unhatted symbols denote artifact
types, whereas the hatted symbols in Equation~\ref{eq:task-formulation} denote a
candidate's realized outputs. Candidate-declared provenance and support pointers
propose relations, which the evaluator verifies against these executed
observations.
Primary claims require both a complete computational path and claim-relevant
visual support with matching population, grouping, quantity, direction, and
required uncertainty.

\subsection{Artifact and Relation Verification}

The evaluator scores graph nodes and edges separately so that an incorrect
artifact can be distinguished from an invalid dependency between otherwise
plausible artifacts. Candidate artifacts are canonicalized and matched to hidden
semantic slots under type and scope constraints. Transform fidelity checks data
views, selectors, operation families, and parameters. Result fidelity checks the
estimand, analysis unit, method family, numerical value, and uncertainty against
independent recomputation. Visual fidelity checks bound arrays, encodings,
groups, units, scales, and uncertainty, while claim fidelity checks direction,
magnitude, population scope, inferential strength, and causal ceiling.

Edge verification follows the same executed evidence. The evaluator tests
whether the observed data produced the transformation, the transformation
produced the result, the rendered marks encode the same quantities, and the
result and marks support the claim. Appendix~\ref{app:protocol-tables},
Table~\ref{tab:edge-verification}, specifies the evidence and pass condition for
every relation. Lineage, numerical recomputation, visual binding, and graph
connectivity are checked programmatically. A frozen semantic matcher handles
residual paraphrase and scope matching for atomic claims but cannot override a
failed structured, numerical, or provenance check.

\subsection{Claim-Path Scoring and Failure Localization}

The hidden specification represents accepted method families, visual encodings,
and claim-support rules as semantic constraints rather than a single gold
program or image. Let $\mathcal{P}_{\mathrm{comp}}(c)$ and
$\mathcal{P}_{\mathrm{vis}}(c)$ denote the accepted computational and visual
support paths for claim $c$, and let each verified node or edge item $z$ have
score $s(z)\in[0,1]$. \ours computes
\begin{equation}
  S_k(c)=\max_{p\in\mathcal{P}_k(c)}\min_{z\in p}s(z),
  \quad k\in\{\mathrm{comp},\mathrm{vis}\},
  \qquad
  S(c)=\min\{S_{\mathrm{comp}}(c),S_{\mathrm{vis}}(c),F_C(c)\},
  \label{eq:claim-support}
\end{equation}
where $F_C(c)$ is local claim fidelity. The inner minimum applies the
weakest-link rule to one path, and the outer maximum admits any complete,
approved alternative. Generated claims are matched one-to-one to required claim
slots, and their verified support mass yields evidence-chain precision, recall,
and F1. Strict evidence-chain success requires execution, complete support for
every required claim, no unsupported primary claim, and satisfaction of all hard
visual constraints. Artifact fidelities, edge pass rates, and an
execution-gated soft chain aggregate provide complementary diagnostic views.

For an incomplete chain, \ours reports the earliest topological layer from
which no accepted partial path can reach the claim: contract/execution,
Data$\to$Transform, Transform$\to$Result, Transform/Result$\to$Visual, or
Result/Visual$\to$Claim. An incomplete path does not determine the score when
another independently verified path remains complete, while downstream
presentation quality cannot repair an upstream scientific error.

\section{\ours Benchmark}
\label{sec:benchmark}

\begin{figure*}[t]
  \centering
  \includegraphics[width=\textwidth]{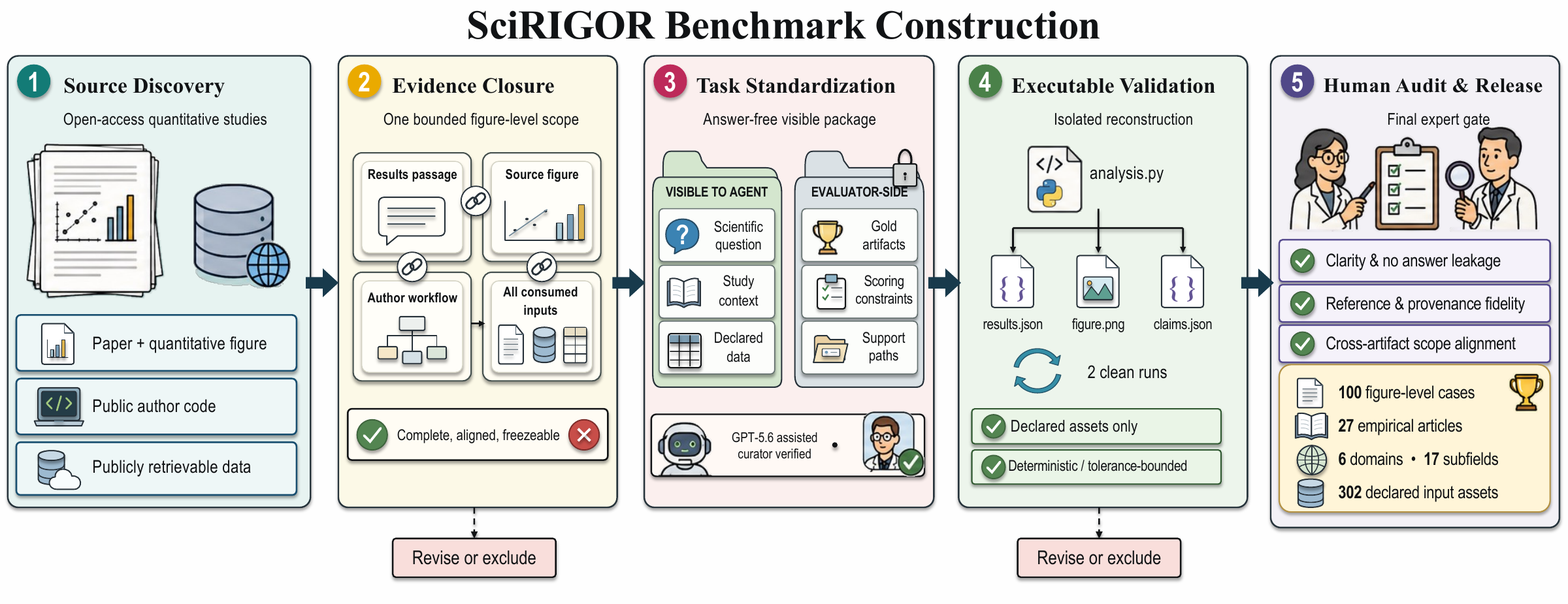}
  \caption{\textbf{\ours benchmark construction.} Open-access studies enter the
  benchmark only after figure-level evidence closure, task standardization, two
  clean reconstructions, and expert audit. The agent-visible task is separated
  from evaluator-side gold artifacts, scoring constraints, and support paths.}
  \label{fig:benchmark-construction}
\end{figure*}

The \ours benchmark instantiates the evaluation framework with 100
evidence-complete, figure-level cases from 27 empirical articles. Each case
aligns a scientific question with all code-consumed inputs, an executable source
analysis, numerical results, a source visualization, atomic findings, and
provenance within one experimental scope. We construct this alignment through
source discovery, evidence closure, standardized task assembly, executable
reproduction, and final human audit (\figref{fig:benchmark-construction}).

\subsection{Source Discovery and Evidence Closure}
\label{sec:source-discovery}

We searched open-access quantitative studies for figures accompanied by public
source data and author-released plotting code. Discovery combined a Europe PMC
search of the \emph{Nature} and \emph{Science} main journals with audited Nature
Portfolio and other open-access sources. For each candidate, automated indexing
connected in-text figure references with full-text passages, supplements, Data
Availability statements, repository records, and exposed source-data links.
These links formed a review queue; admission depended on evidence completeness
rather than publication venue (Appendix~\ref{app:source-discovery-details}).

\label{sec:case-curation}
An admitted case contains four mutually aligned source components: an
author-generated quantitative figure from an auditable paper version; a bounded
Results passage that describes and interprets the selected figure or panel; the
complete author-released script or notebook that generates that target; and
every raw, processed, or intermediate file consumed by its entry point. The
plotting program must have a stable source locator, and every data dependency
must be publicly retrievable and freezeable. Together, these requirements tie
the visual target, experimental interpretation, executable analysis, and inputs
to the same scientific question.

Evidence closure required two clean executions of the frozen author entry point
against the complete inputs. The reproduced quantities, visual marks, groups,
units, and uncertainty encodings had to agree with the paper figure and its
analysis. Each accepted case covers one figure or one coherent panel subset, and
its question, passage, code, inputs, and findings are bounded to that scope. The
visual reference is an original source raster or an audited source-pixel crop.
Missing dependencies, unrecoverable semantics, or substantive disagreement led
to exclusion under the rules detailed in
Appendix~\ref{app:case-curation-details}.

\subsection{Task Construction and Standardization}
\label{sec:case-construction}

Each closed source bundle became a self-contained task without answer-bearing
content. We retained an explicit paper question when it matched the selected
figure scope; otherwise, curators wrote a neutral question bounded to the
experiment. The visible package contains the question and study context in
the task specification together with the declared analysis-ready data. Source
references, gold artifacts, construction metadata, and scoring rules remain
evaluator-side. Appendix~\ref{app:case-construction-details} specifies the
question, visibility, storage, and serialized interface.

Standardization began after source closure. GPT-5.6 assisted with locating
target-specific operations, drafting neutral task descriptions, mapping
heterogeneous tables into documented assets, and proposing source-grounded
Python translations and structured records. Curators verified every proposal
against the frozen author workflow and retained unchanged author code for audit.
Original Python operations were preserved; non-Python workflows were translated
into Python ports that retained source selectors, transformations, estimators,
scales, uncertainty layers, and plot structure. Execution and visual-semantic
comparison validated each port before case admission.

\subsection{Executable Quality Control}
\label{sec:validation-statistics}

Every assembled case ran twice in an isolated environment containing only its
executable reference analysis and declared inputs. Both runs had to terminate
successfully, regenerate the required results, visualization, and claims,
support provenance reconstruction, consume only declared assets, and agree
under a deterministic or explicitly tolerance-bounded contract. Structured
hidden constraints were then checked against the regenerated numerical results, visual
semantics, data bindings, and atomic-claim support. All 100 released cases passed
both clean reconstruction runs.

Curators completed a final audit of question clarity, answer leakage, reference
fidelity, scoring-constraint coherence, and alignment among the selected panel,
source passage, author code, consumed data, and claims. Unresolved scope,
provenance, or support discrepancies triggered revision or exclusion. This
construction-time audit fixes the hidden reference and support rules, enabling
routine model submissions to be scored automatically. Candidate methods outside
the frozen alternatives enter expert adjudication; accepted alternatives are
added to the support specification before rescoring. Appendix~
\ref{app:validation-release-details} provides the complete execution, asset, and
review protocol.

\subsection{Benchmark Composition}

\begin{figure*}[t]
  \centering
  \begin{minipage}[t]{0.65\textwidth}
    \centering
    \includegraphics[width=\linewidth]{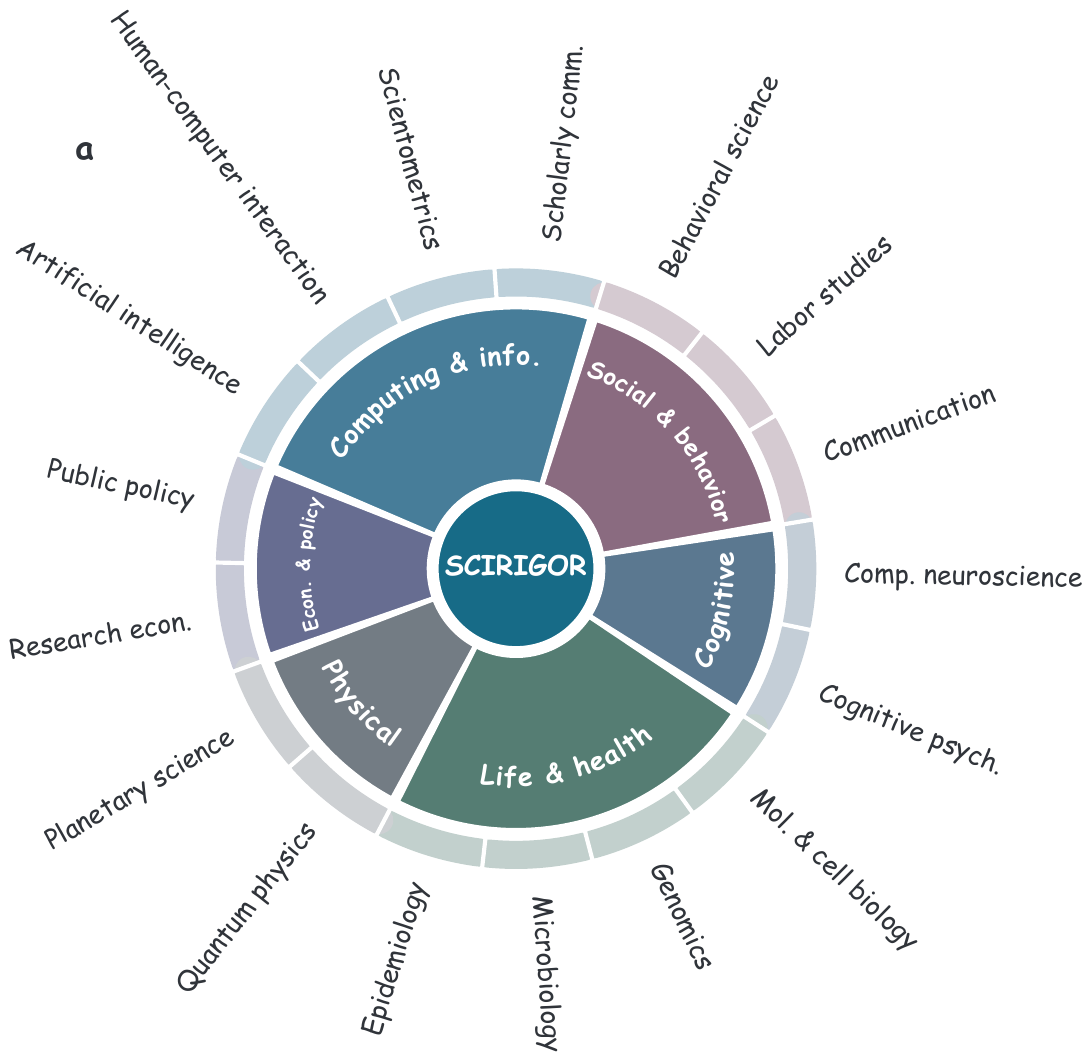}
  \end{minipage}
  \hfill
  \begin{minipage}[t]{0.31\textwidth}
    \centering
    \includegraphics[width=\linewidth]{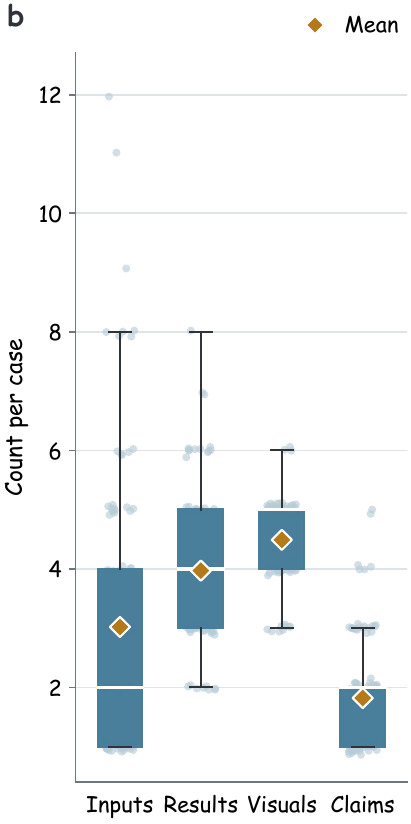}
  \end{minipage}
  \vspace{-0.65em}

  \includegraphics[width=\textwidth]{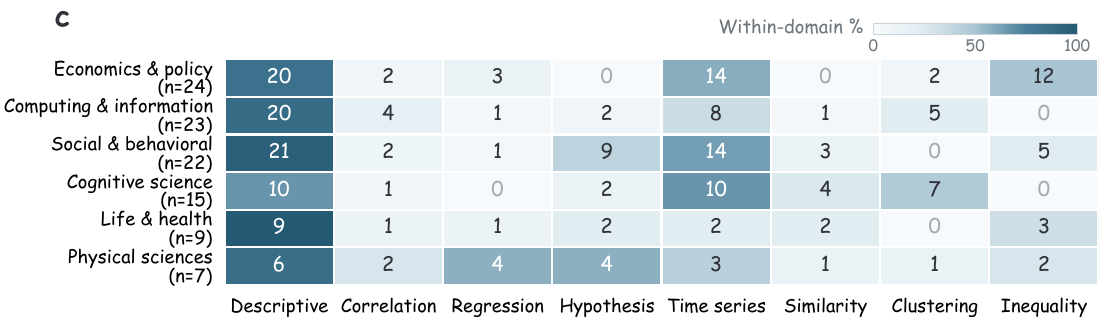}
  \caption{\ours spans diverse scientific domains, analysis methods, and
  multi-artifact task complexities. (a) The 100 cases span six reporting domains
  and 17 disciplinary subfields. The taxonomy ring assigns equal angular width
  to every subfield, while each inner domain spans the subfields it contains.
  (b) Cases require multiple input assets, result slots, visual requirements,
  and atomic claims, creating long evidence chains. (c) Method families are
  multi-label across domains; cells report case counts, and darker color denotes
  greater within-domain prevalence.}
  \label{fig:benchmark-statistics}
\end{figure*}

The release covers six reporting domains defined by the primary scientific
subject of each source study: economics and policy (24 cases); computing and
information sciences (23); social and behavioral sciences (22); cognitive
science (15); life and health sciences (9); and physical sciences (7).
Multiple cases may originate from one article because the benchmark unit is a
bounded figure-level question. The released metadata records the complete
case-to-subfield mapping shown in \figref{fig:benchmark-statistics}a.

The cases vary in the number of input assets, required result slots, visual
requirements, and atomic claims, producing evidence chains of different lengths
(\figref{fig:benchmark-statistics}b). Analysis methods are multi-label:
descriptive aggregation appears across all domains, physical sciences contain
more regression, model-fitting, and inferential checks, cognitive science
contributes trajectory, similarity, and classification analyses, and
economics/policy contains many distribution and inequality computations
(\figref{fig:benchmark-statistics}c). These distributions describe coverage
rather than model performance. They reflect the availability of figure-specific
public code and complete code-consumed data, not a probability-balanced sample
of the scientific literature.

\section{Experiments}
\label{sec:experiments}

\subsection{Main Results}
\label{sec:main-results}

Table~\ref{tab:main-results} evaluates 11 agent/model configurations under the
five-artifact interface. Metrics are percentages over each run's evaluated
cases; execution failures and missing required artifacts receive zero. We report
artifact fidelity, Result$\to$Claim consistency, and evidence-chain composition
to distinguish local correctness from end-to-end scientific support.

\newcommand{\ScoreCell}[2]{\cellcolor{pathblue!#1!white}#2}

\begin{table*}[t]
  \centering
  \caption{Main results on \ours (\%; higher is better). Blue shading is
  normalized within each model row.}
  \label{tab:main-results}
  \small
  \setlength{\tabcolsep}{5.2pt}
  \renewcommand{\arraystretch}{1.12}
  \begin{adjustbox}{max width=\textwidth}
  \begin{tabular}{@{}l *{8}{c}@{}}
    \toprule
    Model
    & Exec.
    & T-Fid.
    & R-Fid.
    & V-Fid.
    & C-Fid.
    & R$\to$C
    & SoftECS
    & Strict \\
    \midrule
    GPT-5.5 (xhigh)
    & \ScoreCell{40}{100.0}
    & \ScoreCell{19}{57.9} & \ScoreCell{24}{68.2} & \ScoreCell{27}{72.8} & \ScoreCell{19}{57.2}
    & \ScoreCell{37}{93.4} & \ScoreCell{22}{62.6} & \ScoreCell{0}{18.0} \\
    GPT-5.5 (high)
    & \ScoreCell{40}{100.0}
    & \ScoreCell{20}{58.5} & \ScoreCell{26}{69.8} & \ScoreCell{26}{70.8} & \ScoreCell{21}{59.1}
    & \ScoreCell{36}{91.3} & \ScoreCell{21}{59.2} & \ScoreCell{0}{16.0} \\
    GPT-5.6-Sol (high)
    & \ScoreCell{40}{100.0}
    & \ScoreCell{21}{59.5} & \ScoreCell{24}{64.7} & \ScoreCell{27}{72.8} & \ScoreCell{21}{58.7}
    & \ScoreCell{35}{90.0} & \ScoreCell{22}{61.1} & \ScoreCell{0}{13.0} \\
    GPT-5.6-Sol (xhigh)
    & \ScoreCell{40}{100.0}
    & \ScoreCell{21}{58.7} & \ScoreCell{24}{64.8} & \ScoreCell{27}{70.9} & \ScoreCell{21}{57.7}
    & \ScoreCell{35}{88.9} & \ScoreCell{21}{57.6} & \ScoreCell{0}{11.0} \\
    GPT-5.5 (low)
    & \ScoreCell{25}{63.0}
    & \ScoreCell{23}{58.3} & \ScoreCell{20}{50.4} & \ScoreCell{31}{75.4} & \ScoreCell{18}{46.4}
    & \ScoreCell{40}{97.8} & \ScoreCell{11}{28.0} & \ScoreCell{0}{3.0} \\
    Qwen3.8-Max
    & \ScoreCell{40}{47.0}
    & \ScoreCell{20}{24.1} & \ScoreCell{14}{18.2} & \ScoreCell{28}{34.0} & \ScoreCell{19}{23.5}
    & \ScoreCell{40}{46.5} & \ScoreCell{16}{20.1} & \ScoreCell{0}{2.0} \\
    Kimi-K3
    & \ScoreCell{40}{79.0}
    & \ScoreCell{22}{43.5} & \ScoreCell{18}{36.4} & \ScoreCell{28}{55.9} & \ScoreCell{20}{40.2}
    & \ScoreCell{37}{73.2} & \ScoreCell{17}{35.0} & \ScoreCell{0}{1.0} \\
    GPT-5.6 (low)
    & \ScoreCell{36}{88.0}
    & \ScoreCell{23}{55.5} & \ScoreCell{19}{47.2} & \ScoreCell{32}{77.3} & \ScoreCell{21}{52.5}
    & \ScoreCell{40}{97.9} & \ScoreCell{18}{43.0} & \ScoreCell{0}{0.0} \\
    Claude Sonnet 5
    & \ScoreCell{40}{75.0}
    & \ScoreCell{24}{44.5} & \ScoreCell{22}{40.6} & \ScoreCell{29}{54.3} & \ScoreCell{13}{24.0}
    & \ScoreCell{38}{71.9} & \ScoreCell{12}{22.1} & \ScoreCell{0}{0.0} \\
    Qwen3-Coder-30B-A3B
    & \ScoreCell{40}{29.0}
    & \ScoreCell{13}{9.3} & \ScoreCell{11}{7.8} & \ScoreCell{24}{17.6} & \ScoreCell{7}{5.0}
    & \ScoreCell{29}{21.1} & \ScoreCell{2}{1.7} & \ScoreCell{0}{0.0} \\
    Qwen-AgentWorld-35B-A3B
    & \ScoreCell{40}{12.0}
    & \ScoreCell{17}{5.1} & \ScoreCell{11}{3.4} & \ScoreCell{24}{7.1} & \ScoreCell{20}{6.1}
    & \ScoreCell{38}{11.4} & \ScoreCell{0}{0.0} & \ScoreCell{0}{0.0} \\
    \bottomrule
  \end{tabular}
  \end{adjustbox}
\end{table*}

\paragraph{Execution remains a substantive bottleneck.}
Execution gates every downstream scientific score. Kimi-K3 executes 79.0\% of
its assigned cases, Qwen3.8-Max executes 47.0\%, Qwen3-Coder-30B-A3B reaches
29.0\%, and Qwen-AgentWorld-35B-A3B reaches 12.0\%. Scores computed only over
successful outputs would therefore conflate scientific quality with selective
completion.

\paragraph{Candidate self-consistency can mask scientific error.}
Across all 11 rows, R$\to$C is the highest or second-highest displayed score,
while result and claim fidelity are substantially lower for several
high-execution runs. GPT-5.6 (low), for example, obtains 97.9\%
Result$\to$Claim consistency but only 47.2\% result fidelity and 52.5\% claim
fidelity. Its claims therefore tend to agree with its own reported results even
when those results depart from the scientific target. Candidate
self-consistency is necessary for a coherent submission but is not sufficient
evidence of scientific correctness.

\paragraph{Partial success collapses under chain composition.}
The strongest run reaches 62.6\% SoftECS but only 18.0\% Strict success. Kimi-K3
falls from 79.0\% execution to 35.0\% SoftECS and 1.0\% Strict success. These
gaps show that disconnected artifact averages are insufficient: a submission can
accumulate several locally successful components without completing a single
fully supported data-to-claim path. Fidelity and edge consistency should be
reported separately before applying an end-to-end chain gate.

\subsection{Error Analysis}
\label{sec:error-analysis}

\begin{figure*}[t]
  \centering
  \includegraphics[width=\textwidth]{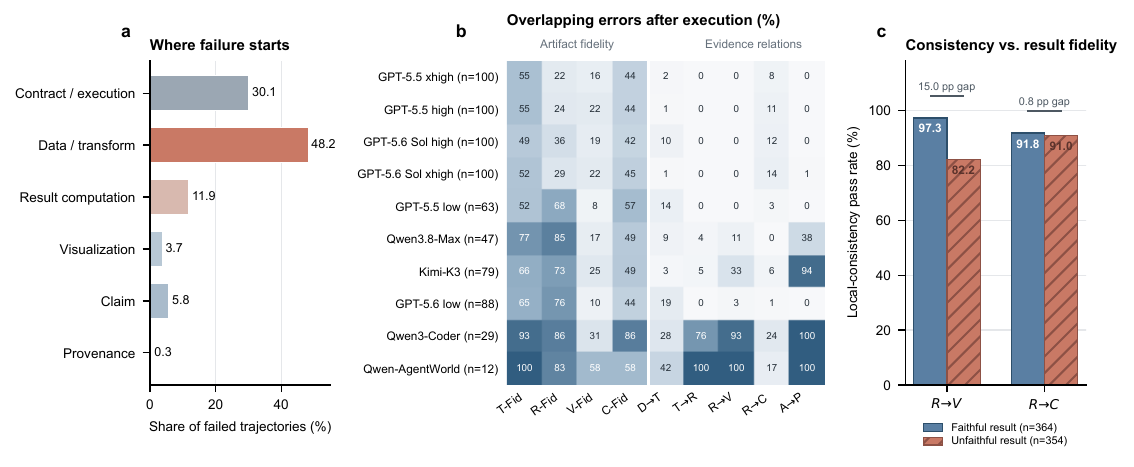}
  \caption{Case-level error analysis across ten complete 100-case runs.
  (a) Earliest failure stage among the 936 trajectories that do not complete a
  strict chain. Each trajectory contributes to exactly one stage.
  (b) Overlapping artifact- and edge-level errors among the 718 successfully
  executed trajectories; values are percentages within each model's executed
  cases. (c) Local Result$\to$Visual and Result$\to$Claim consistency conditioned
  on independent result fidelity.}
  \label{fig:error-analysis}
\end{figure*}

We diagnose failures from three complementary views: the earliest broken stage,
the full set of artifact and edge errors after successful execution, and whether
local consistency distinguishes faithful from unfaithful results. These labels
are produced by the evaluator and may overlap in
\figref{fig:error-analysis}b; they describe observable failure signatures rather
than assigning a unique causal explanation to a trajectory.

\paragraph{Most chains break before a faithful result is established.}
Among 936 non-strict trajectories, 48.2\% first fail in data selection or
transformation and 30.1\% fail at the artifact contract or execution gate
(\figref{fig:error-analysis}a). Result computation accounts for a further
11.9\%, whereas visualization, claims, and provenance together account for only
9.8\% of first failures. Thus, late-stage errors are visible, but the dominant
bottleneck is establishing a valid computational basis for downstream evidence.

\paragraph{Scientific errors persist after execution succeeds.}
Conditioning on the 718 executed trajectories removes mechanical non-completion
from the comparison. Transform fidelity fails in 59.6\% of these trajectories,
result fidelity in 49.3\%, and claim fidelity in 47.9\%
(\figref{fig:error-analysis}b). By contrast, Result$\to$Claim consistency fails
in only 8.6\%. The model-level heatmap shows that this gap is not confined to a
single run: systems frequently propagate their own intermediate values
coherently even when those values do not match the scientific target.

\paragraph{Local self-consistency is nearly invariant to result correctness.}
Result$\to$Claim consistency passes on 91.8\% of result-faithful trajectories
and 91.0\% of result-unfaithful trajectories
(\figref{fig:error-analysis}c). Result$\to$Visual consistency is more sensitive,
but still passes on 82.2\% of trajectories with unfaithful results, compared
with 97.3\% when results are faithful. A candidate can therefore explain and
visualize its own result without establishing that the result is correct. This
case-level conditional analysis confirms that cross-artifact coherence is
necessary but cannot replace independent fidelity checks.

\subsection{Complexity Sensitivity}
\label{sec:complexity-sensitivity}

\begin{figure*}[t]
  \centering
  \includegraphics[width=\textwidth]{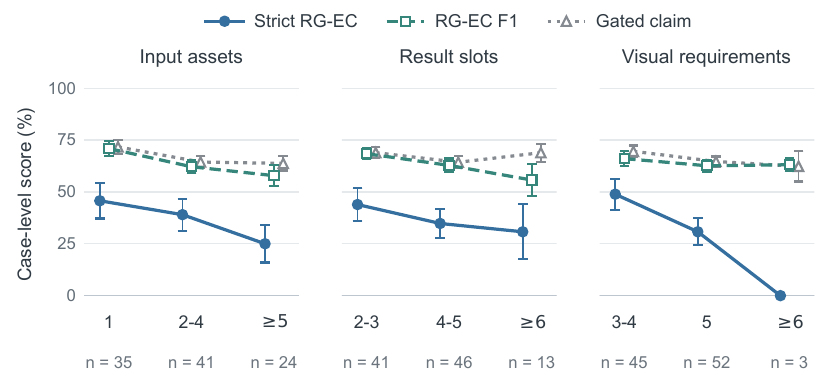}
  \caption{Performance by task complexity for GPT-5.6 Sol xhigh on 100 cases.
  Points show bin means; error bars denote $\pm 1$ SEM across cases.
  Strict RG-EC decreases with more input assets, required
  result slots, and visual requirements; RG-EC F1 also declines with assets and
  result slots, while the gated-claim composite is flatter. RG-EC F1 gates claim
  coverage by verified numerical support, whereas Strict RG-EC requires complete
  support for every required and generated claim.}
  \label{fig:complexity-performance-sensitivity}
\end{figure*}

\figref{fig:complexity-performance-sensitivity} shows that complexity primarily
affects chain completeness rather than surface claim plausibility. Moving from
one input asset to at least five lowers Strict RG-EC from 45.7\% to 25.0\% and
RG-EC F1 from 70.9\% to 57.8\%, whereas the gated-claim score drops from 71.7\%
to 63.7\%. Required result slots show the same pattern: Strict RG-EC decreases
from 43.9\% to 30.8\% and RG-EC F1 from 68.4\% to 55.7\%, while the gated-claim
score remains nearly unchanged. This divergence supports scoring
result-grounded claim paths rather than final claims alone. The highest
visual-requirement bin contains only three cases and is therefore a stress-test
signal, not a standalone estimate.

\subsection{Evaluation Scope}

The experiments score observable artifacts submitted through the five-file
interface. The evaluator reconstructs their semantic relations but does not
observe latent reasoning or agent action trajectories. The results therefore
compare artifact fidelity and complete evidence paths and localize observable
breaks; they do not identify whether a failure arose from planning,
implementation, or unsupported generation.

\section{Conclusion}

\ours formulates evidence-grounded scientific analysis with multimodal artifacts
around a verified data-to-claim support path. Its evaluation framework separates artifact
fidelity from cross-artifact consistency, reconstructs computational and visual
dependencies, accepts source-grounded alternative paths, and localizes the
earliest unsupported relation. The benchmark's 100 evidence-complete,
figure-level cases
align questions, inputs, source analyses, results, visualizations, findings, and
provenance within one audited scope. Across 11 agent/model configurations, high
execution or
Result$\to$Claim consistency can coexist with much lower target fidelity, and
strict chain completion remains rare. Across the ten complete 100-case runs,
78.3\% of failed trajectories first break at the contract/execution or
data/transformation stage; among executed trajectories, Result$\to$Claim
consistency passes at nearly the same rate for faithful and unfaithful results
(91.8\% versus 91.0\%).

Together, these results support the paper's central evaluation principle:
execution, visual plausibility, final-output similarity, and candidate
self-consistency are not substitutes for independently verified relations from
supplied data to each reported claim.

\clearpage
\bibliography{reference}
\bibliographystyle{iclr2027_conference}
\clearpage
\appendix
\section{Executable Contract and Edge-Verification Details}
\label{app:protocol-tables}

The main paper presents the task interface and evidence-graph workflow at a
conceptual level. Tables~\ref{tab:five-file-contract} and
\ref{tab:edge-verification} provide the complete artifact ownership and
programmatic edge-verification rules used by the evaluator.

\begin{table*}[t]
\centering
\caption{Five-file \ours contract for reference and candidate bundles.}
\label{tab:five-file-contract}
\footnotesize
\setlength{\tabcolsep}{4pt}
\renewcommand{\arraystretch}{1.12}
\begin{tabularx}{\textwidth}{@{}>{\raggedright\arraybackslash}p{0.20\textwidth}>{\raggedright\arraybackslash}X>{\raggedright\arraybackslash}X@{}}
\toprule
\textbf{Serialized artifact} & \textbf{Frozen reference role} & \textbf{Candidate execution role} \\
\midrule
\texttt{analysis.py} ($T$) & Audited executable reconstruction of the gold analysis. & Model-written analysis and plotting entry point. \\
\texttt{results.json} ($R$) & Recomputed gold numerical results. & Model-computed numerical results. \\
\texttt{figure.png} ($V$) & Original paper figure or bounded source-pixel crop. & Data-generated visualization. \\
\texttt{claims.json} ($C$) & Data-verifiable atomic findings from the source analysis. & Model-generated atomic findings. \\
\texttt{minimal\_}\allowbreak\texttt{provenance.json} ($P$) & Gold source and code-location hints. & Evaluator-verified input, execution, and code-location provenance. \\
\bottomrule
\end{tabularx}
\end{table*}

\begin{table*}[t]
\centering
\caption{Independent evidence used to verify each candidate relation. Relation
symbols follow Equation~\ref{eq:evidence-graph}; both incident nodes must satisfy
their semantic constraints before an edge can pass.}
\label{tab:edge-verification}
\footnotesize
\setlength{\tabcolsep}{4.5pt}
\renewcommand{\arraystretch}{1.12}
\begin{tabularx}{\textwidth}{@{}>{\raggedright\arraybackslash}p{0.10\textwidth}>{\raggedright\arraybackslash}p{0.35\textwidth}>{\raggedright\arraybackslash}X@{}}
\toprule
\textbf{Relation} & \textbf{Evaluator evidence} & \textbf{Pass condition} \\
\midrule
$D\to T$
& Sandbox file-access trace, selected row/column identifiers, operation
parameters, and input/output fingerprints
& The executed semantic data view and transformation agree with an allowed
source view and operation family. \\
$T\to R$
& Captured transform output and an independent case-specific numerical oracle
& The estimand, analysis unit, method family, value, and required uncertainty
recompute within the declared tolerance. \\
$T/R\to V$
& Captured plotting arrays, artist bindings, and reconstructed Figure IR
& Visual marks encode the executed quantities with the required population,
grouping, channels, units, scales, and uncertainty. \\
$R/V\to C$
& Structured claim fields, verified results and marks, and the hidden claim
envelope
& Direction, magnitude, population, uncertainty, inferential strength, and
causal scope are entailed by at least one approved computational and visual path.
\\
\bottomrule
\end{tabularx}
\end{table*}
\FloatBarrier

\section{Source Discovery and Figure-Level Indexing Details}
\label{app:source-discovery-details}

\paragraph{Search queries and source verification.} One discovery stream
queried the \emph{Nature} and \emph{Science} main journals through Europe PMC,
combining journal ISSNs with open-access, full-text, research-article, and CC BY
filters. We rechecked returned records for journal identity, article type,
license, and source version. The ISSN constraint prevented expansion to other
journals in the same publisher family. This crawler supplied candidates rather
than defining venue eligibility: audited Nature Portfolio and other open-access
sources faced the same evidence-completeness criteria.

\paragraph{Figure-level indexing.} For each retrieved article, the pipeline
collected full-text XML, source-native figures, machine-readable supplements,
and exposed source-data links. It resolved in-text figure references and indexed
the associated body paragraphs, quantitative statements, Methods sections, Data
Availability statements, and repository identifiers. These signals constructed
a review queue only; a detected figure--data link did not establish availability
of target plotting code or all code-consumed inputs.

\section{Evidence-Complete Curation Details}
\label{app:case-curation-details}

\paragraph{Component-level exclusions.} Schematics, decorative illustrations, and image-only panels without recoverable data bindings did not satisfy the figure requirement. Captions, headings, and nearby discussion of other experiments could not replace a bounded Results passage. General software, upstream preprocessing alone, code for another panel, and curator-written plotting code could not replace the original target plotting program, which required a stable locator such as a commit, file, function, line range, or notebook cell. A publisher summary table was insufficient when the target code also consumed unavailable raw data, mappings, model outputs, checkpoints, or configurations.

\paragraph{Reproduction audit.} Reviewers ran the frozen author entry point
twice in a pinned environment against complete frozen inputs and compared the
reproduced quantities, marks, groups, units, and uncertainty encodings with the
source figure and analysis. Private-path dependencies, unrecoverable semantics,
or substantive disagreement with the published result caused rejection. The
source registry retains the original source, environment, file-open traces,
execution logs, and comparison report. Independent reconstruction could
document a rejection but could not substitute for missing author code.

\paragraph{Panel and crop audit.} We retained the full paper and source figure.
When a case covered only part of a composite figure, its visual reference was a
logged, unresized crop of the selected source panels. Recorded crop coordinates,
dimensions, and hashes preserve the link to the source image; redraws were not
accepted as references.

\section{Standardized Case Construction Details}
\label{app:case-construction-details}

\paragraph{Question checks and visibility boundaries.} Curators checked every
question for leading language, answer-bearing values, vague comparison sets,
and undefined analysis units. The visible package contains only
\texttt{input/task.json}, declared assets under \texttt{input/data/}, and the
scientific context and measurement semantics needed to solve the task. The
source paper, unchanged author code, original figure, gold findings, scoring
rules, and construction metadata remain evaluator-side.

\paragraph{Evaluator-side storage.} The \texttt{source\_registry/} retains the
source PDF, unchanged author code, data hierarchy, reproduction output, crop
record, and execution logs; \texttt{hidden\_evaluation/} stores source views and
scoring constraints. Candidate figures are checked for marks, field bindings,
axes, units, uncertainty channels, and panel semantics rather than pixel
identity with the frozen paper raster.

\paragraph{GPT-5.6-assisted curation.} Where applicable, GPT-5.6 helped locate
target-specific operations, draft neutral task descriptions, map heterogeneous
tables into documented visible assets, and propose source-grounded Python
translations and structured records. Curators checked every proposal against
the frozen author workflow. Only source-present code, data, figures, and claims
could enter an accepted case.

\paragraph{Python-port validation.} Original Python operations were retained.
R, R Markdown, and other workflows were translated while preserving source
selectors, transformations, estimators, scales, uncertainty layers, and plot
structure. Execution and visual-semantic comparison validated each port, with
library- or platform-sensitive differences recorded explicitly. Every
\texttt{analysis.py} used only visible inputs, made no network calls, and
regenerated \texttt{results.json}, \texttt{figure.png}, and
\texttt{claims.json} without copying frozen references; the evaluator verified
provenance separately.

\section{Validation and Release Details}
\label{app:validation-release-details}

\paragraph{Asset integrity and execution environment.} The release contains 302
declared input assets, each materialized at its specified path, matched to its
recorded SHA-256 hash, and stored as a regular file rather than a symbolic link.
File-open traces confirmed that execution consumed only declared inputs. One
numerically sensitive case used its pinned SciPy environment; all 100 cases
passed two clean reconstruction runs.

\paragraph{Final human-review gate.} Final curator approval was required for
release. The audit assessed the visible task for ambiguity and
answer leakage, verified a shared scientific scope across the selected panel,
experimental passage, author code, consumed data, reference artifacts, and
atomic claims, and checked hidden constraints for unsupported or contradictory
requirements. Unresolved discrepancies triggered revision or rejection. All 100
released cases record approved review status.

\paragraph{Disciplinary assignment.}
Each case belongs to one reporting domain defined by the primary scientific
system or institutional setting in the source paper, not by statistical method
or visualization type. Table~\ref{tab:disciplinary-composition} summarizes the
resulting composition.

\begin{table*}[t]
\centering
\caption{Disciplinary composition of the 100 \ours cases. Domains are used for descriptive reporting and do not imply a probability-balanced sample of the scientific literature.}
\label{tab:disciplinary-composition}
\footnotesize
\setlength{\tabcolsep}{5pt}
\renewcommand{\arraystretch}{1.12}
\begin{tabularx}{\textwidth}{@{}p{0.28\textwidth}Xr@{}}
\toprule
\textbf{Reporting domain} & \textbf{Included research topics} & \textbf{Cases} \\
\midrule
Economics and policy
& Research funding, award costs, funding inequality, peer review, scientific careers, and research-workforce composition
& 24 \\

Computing and information sciences
& Language-model reliability, human--computer interaction, preprint adoption, open-access pathways, bibliometrics, and scholarly communication
& 23 \\

Social and behavioral sciences
& Media representation, misinformation, platform labor, urban heat responses, collective behavior, and resource-allocation experiments
& 22 \\

Cognitive science
& Computational cognitive models, hippocampal learning, recognition, categorization, and human behavioral experiments
& 15 \\

Life and health sciences
& Cellular and molecular biology, genetics, microbiology, cancer biology, and influenza surveillance
& 9 \\

Physical sciences
& Atom interferometry, quantum sensing, parity readout, and post-impact asteroid photometry
& 7 \\
\midrule
\textbf{Total} & & \textbf{100} \\
\bottomrule
\end{tabularx}
\end{table*}

\section{Case Studies}
\label{app:case-studies}
We present six representative cases spanning the benchmark's scientific
coverage. Following the compact format of an illustrative task
case, each example specifies the scientific task and frozen evidence, then
summarizes the corresponding executable evidence chain. The displayed
images are frozen paper figures or source-pixel panel crops used by the
evaluator rather than candidate-generated visualizations. They illustrate the
public benchmark contract and are not drawn from the hidden evaluation set.

\subsection{Economics and Policy: NIH Funding Inequality}
\label{app:case-nih-theil}

\paragraph{Task and frozen evidence.}
This case asks how much of the inequality in NIH Research Project Grant
funding is attributable to within-group and between-group differences when
principal investigators are grouped by career stage, gender, race, or degree.
The visible input contains the complete author data and four annual
decomposition tables spanning fiscal years 1985--2020. Each table contains the
total Theil index and its within- and between-group components, giving 144
group--year rows and 432 displayed point instances.

\begin{figure*}[t]
    \centering
    \includegraphics[width=0.74\textwidth]{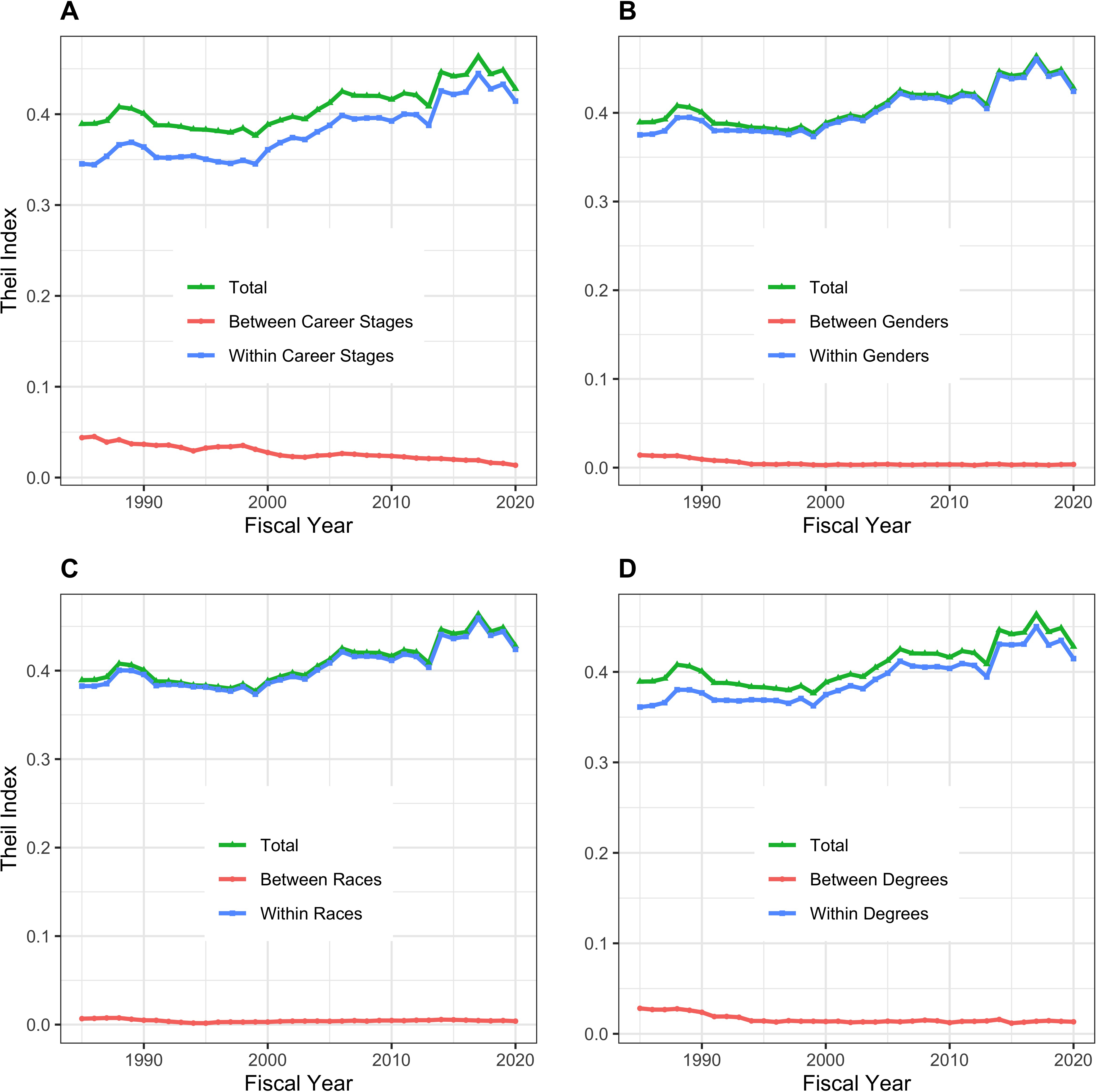}
    \caption{\textbf{NIH funding-inequality case.} The frozen Figure~4
    reference decomposes total Theil inequality into between- and within-group
    components by career stage, gender, race, and degree from FY1985 to FY2020.
    The within-group component exceeds the between-group component throughout
    all four panels.}
    \label{fig:case-nih-theil}
\end{figure*}

\paragraph{Reference evidence chain.}
The executable analysis verifies, for every year and grouping, that the two
components sum to the total Theil index, preserves the author's year and group
selectors, and reconstructs the four-panel trajectory. As shown in
Figure~\ref{fig:case-nih-theil}, within-group differences contribute more to
funding inequality than between-group differences for all four groupings. This
source-grounded finding is an accounting decomposition for funded RPG
principal investigators, not a causal effect or a statement about applicants
or the wider scientific workforce.

\FloatBarrier

\subsection{Social and Behavioral Sciences: Heat and Food-Delivery Workload}
\label{app:case-food-delivery}

\paragraph{Task and frozen evidence.}
This case examines how completed orders per food-delivery worker vary with
daily maximum temperature, with particular attention to the lunch peak above
$35\,^{\circ}\mathrm{C}$. The three visible author tables contain processed
regression outputs for all-day, lunch-peak, and dinner-peak workload across 21
temperature bins. Coefficients and both confidence bounds are on the log scale,
so each must be transformed independently as $100[\exp(x)-1]$ relative to the
$20\,^{\circ}\mathrm{C}$ baseline.

\begin{figure*}[t]
    \centering
    \includegraphics[width=0.72\textwidth]{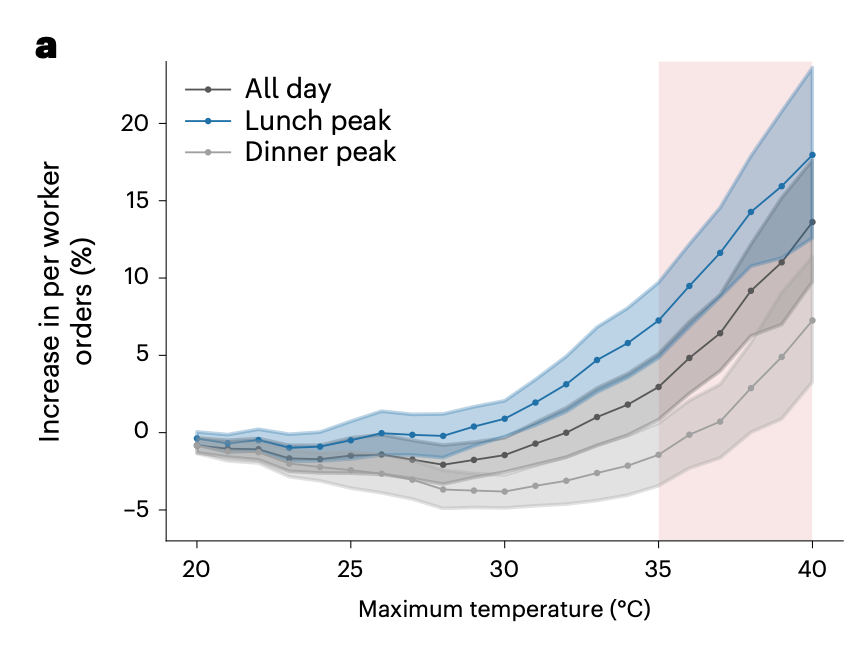}
    \caption{\textbf{Food-delivery workload case.} The frozen Figure~4a panel
    shows percentage changes in completed orders per worker relative to the
    $20\,^{\circ}\mathrm{C}$ baseline for all-day, lunch-peak, and dinner-peak
    periods. Curves and bands encode transformed estimates and 95\% confidence
    intervals; shading marks the extreme-heat range.}
    \label{fig:case-food-delivery}
\end{figure*}

\paragraph{Reference evidence chain.}
The normalized executable retains the author's term ordering, applies the
nonlinear transformation to all 63 estimates and confidence limits, and
reconstructs the response curves and uncertainty bands. As shown in
Figure~\ref{fig:case-food-delivery}, the bounded source analysis reports that
lunch-peak workload is 7.3\% higher in the
$(35,36]\,^{\circ}\mathrm{C}$ bin (95\% CI 4.8--9.7\%) and 18.0\% higher in the
open-ended $(40,\infty)\,^{\circ}\mathrm{C}$ bin (95\% CI 12.6--23.6\%). Because
the inputs are processed model outputs rather than worker-level observations,
the case supports these reported associations but no new causal claim.

\FloatBarrier

\subsection{Computing and Information Sciences: Language-Model Reliability}
\label{app:case-llm-reliability}

\paragraph{Task and frozen evidence.}
This case asks how raw and shaped-up GPT, LLaMA, and BLOOM systems differ in
correctness, robustness to prompt variation, concordance with human difficulty,
and error-free behavior. Three author-released tables provide six normalized
indicators for 32 named systems: 10 GPT, 10 LLaMA, and 12 BLOOM variants. The
family sizes and raw-to-shaped composition are unequal, so pooling all systems
would allow the largest family to dominate the comparison.

\begin{figure*}[t]
    \centering
    \includegraphics[width=0.96\textwidth]{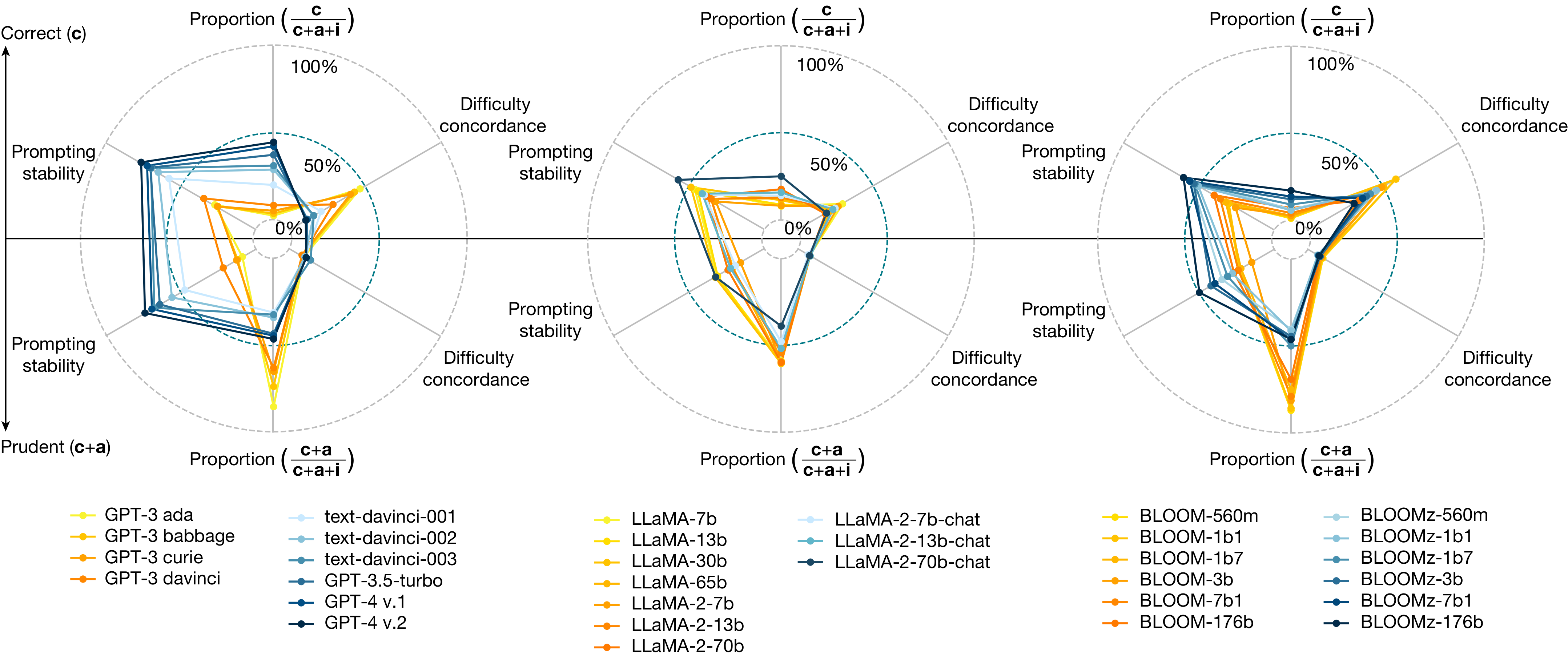}
    \caption{\textbf{Language-model reliability case.} The frozen Figure~1
    reference presents six-indicator profiles for GPT, LLaMA, and BLOOM
    systems on a common radial scale. Warm profiles denote raw systems and blue
    profiles shaped systems, exposing the joint correctness, robustness,
    difficulty-concordance, and error-free trade-offs evaluated by the case.}
    \label{fig:case-llm-reliability}
\end{figure*}

\paragraph{Reference evidence chain.}
The executable analysis first computes raw-versus-shaped summaries within each
family and then averages the three family contrasts with equal family weight.
It also reconstructs the common-scale radar panels from the complete system
profiles. Figure~\ref{fig:case-llm-reliability} visualizes the source finding:
shaped systems are more correct and more stable to prompt variation, but their
correctness is less concordant with human difficulty and they exhibit more
overall failures. The comparison is descriptive for this fixed system set;
the aggregate tables contain no item-level replication or uncertainty with
which to identify a causal effect of post-training.

\FloatBarrier

\subsection{Computing and Information Sciences: bioRxiv Publication Outcomes}
\label{app:case-biorxiv}

\paragraph{Task and frozen evidence.}
This case asks what share of bioRxiv preprints was linked to a journal
publication, which subject categories led by publication rate and count, and
how the observed rate varied with posting date. The two complete author tables
contain 61 monthly cohorts from November 2013 through November 2018 and 27
subject categories. Together they account for 37,648 preprints and 15,797
linked publications, with no imputation, smoothing, or model fitting required.

\begin{figure*}[t]
    \centering
    \includegraphics[width=0.72\textwidth]{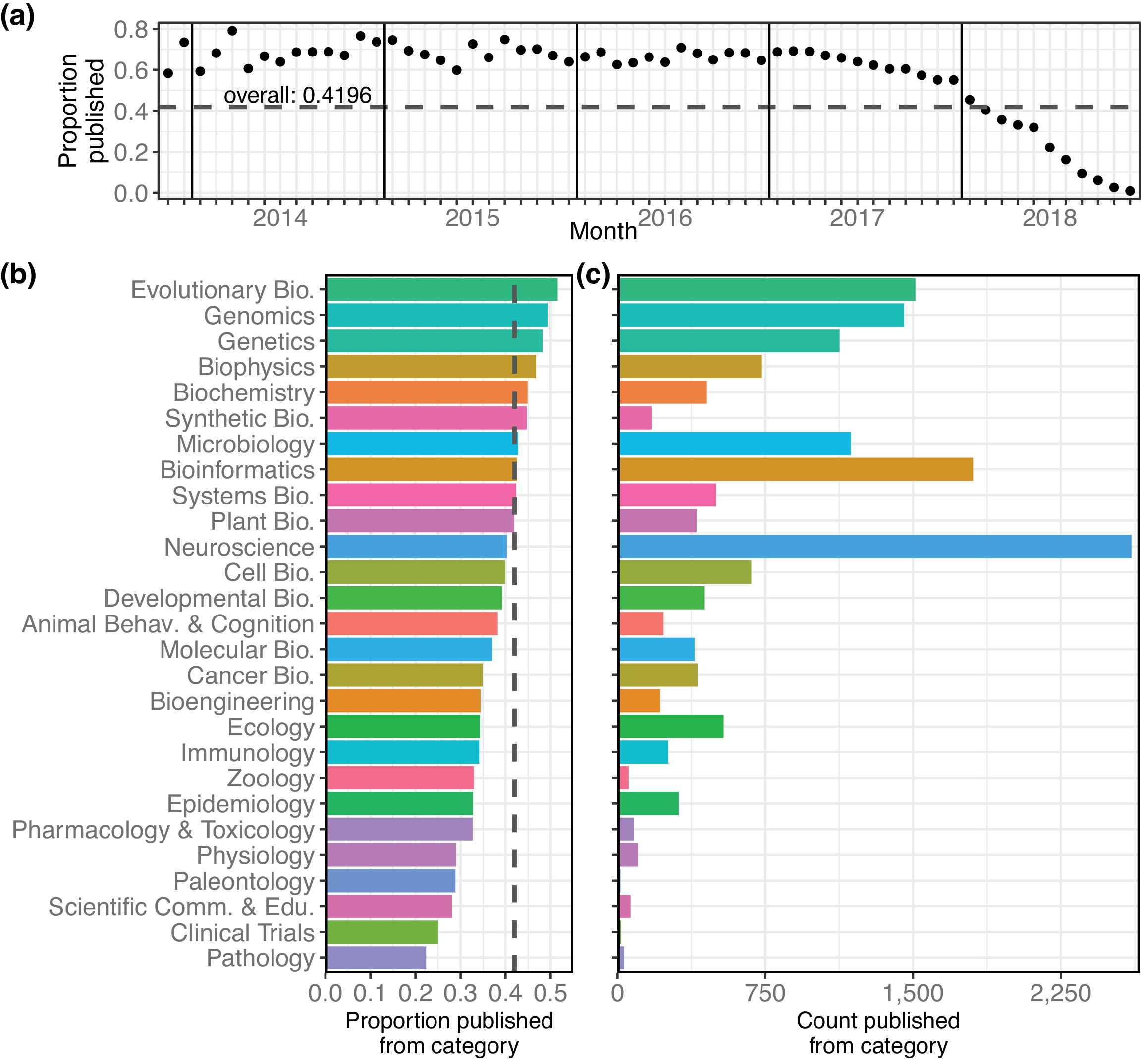}
    \caption{\textbf{bioRxiv publication-outcomes case.} The frozen Figure~3
    reference combines monthly linked-publication proportions with category-
    level proportions and counts. Evolutionary biology has the highest linked
    proportion, whereas neuroscience has the largest linked count; the monthly
    view must be interpreted under unequal follow-up across posting cohorts.}
    \label{fig:case-biorxiv}
\end{figure*}

\paragraph{Reference evidence chain.}
The executable analysis recomputes every row-level proportion, aggregates the
snapshot totals, and independently ranks categories by proportion and count.
As summarized in Figure~\ref{fig:case-biorxiv}, the paper reports an overall
linked-publication rate of 42.0\%, the highest
proportional rate for evolutionary biology (51.5\%), and the largest linked
count for neuroscience (2,608). It also describes rates near zero for the most
recent preprints followed by an increase over roughly 12--18 months. The last
pattern is explicitly snapshot-bound: recent cohorts have shorter follow-up,
so their observed rate is not their eventual publication probability.

\FloatBarrier

\subsection{Life and Health Sciences: Phenotype Distributions across Gleason Groups}
\label{app:case-tcga-gleason}

\paragraph{Task and frozen evidence.}
This case examines how Proliferation and Apoptosis phenotype scores differ
across low, intermediate, and high Gleason groups and how the corresponding
group centroids relate to the model's phenotype directions. The frozen inputs
contain 313 patient-specific Boolean models (23 low, 166 intermediate, and 124
high), the five PCA phenotype coordinates, and the explained variance of each
component. The phenotype scores are semi-quantitative model probabilities, not
direct measurements of cellular rates.

\begin{figure*}[t]
    \centering
    \includegraphics[width=0.88\textwidth]{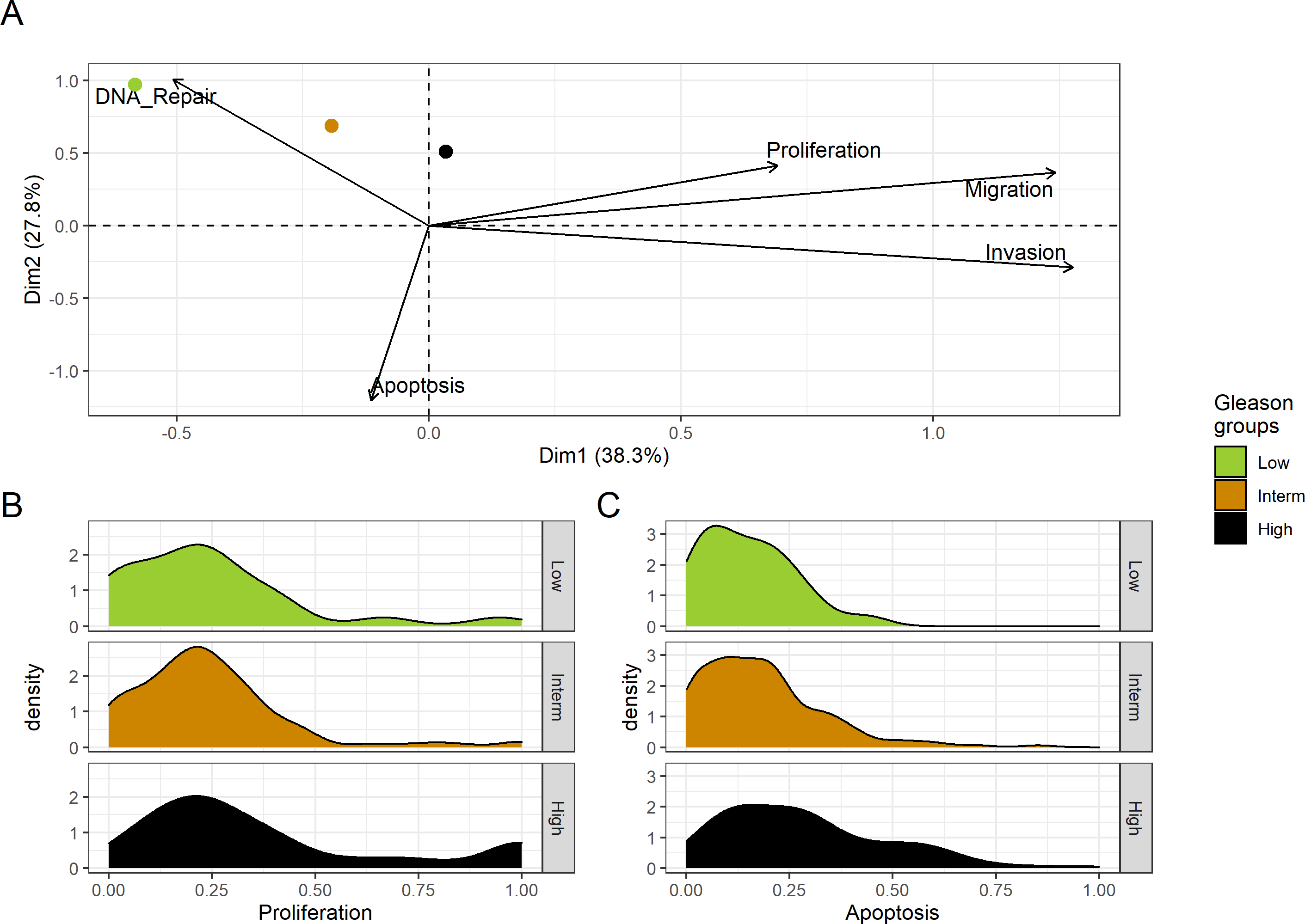}
    \caption{\textbf{Gleason phenotype case.} The frozen Figure~4 reference
    combines PCA phenotype directions and group centroids with complete
    Proliferation and Apoptosis distributions for the low, intermediate, and
    high Gleason groups. The two displayed dimensions explain 38.3\% and
    27.8\% of the model variance.}
    \label{fig:case-tcga-gleason}
\end{figure*}

\paragraph{Reference evidence chain.}
The executable analysis preserves all patient rows, performs tie-corrected
Kruskal--Wallis comparisons, and reconstructs both the PCA panel and the full
group-wise score distributions shown in Figure~\ref{fig:case-tcga-gleason}.
Proliferation differs across groups
($H=12.36$, $p=0.00207$), with a larger high-score tail in the high-Gleason
group. Apoptosis also differs ($H=25.55$, $p=2.83\times10^{-6}$), but the source
analysis states that it shows no clear grade-wise trend. The PCA centroid audit
provides secondary visual context; none of these descriptive comparisons
establishes a causal disease mechanism or treatment effect.

\FloatBarrier

\subsection{Physical Sciences: Post-impact Fading of Didymos}
\label{app:case-dart}

\paragraph{Task and frozen evidence.}
This case asks for Didymos's post-impact fading rate, the time at which the
fitted trend returns to the pre-impact absolute magnitude, and the associated
projected dust velocity. One table contains all 27 plotted measurements and
their observation-level uncertainties from 15 observers; a second identifies
the author's 21-observation fit subset. Visible reference constants specify the
pre-impact magnitude, photometric aperture, pixel scale, and Earth distance.

\begin{figure*}[t]
    \centering
    \includegraphics[width=0.92\textwidth]{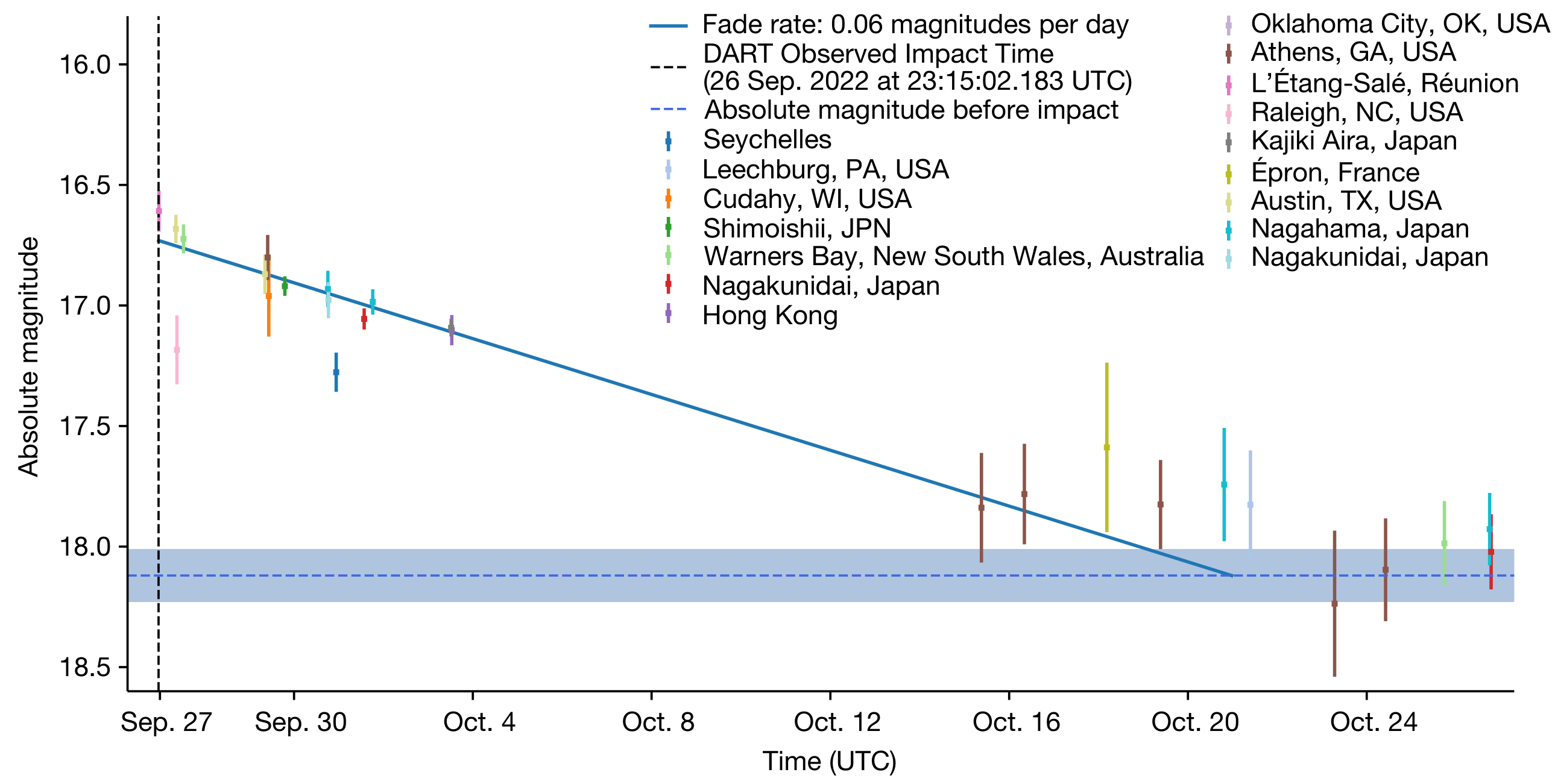}
    \caption{\textbf{Didymos post-impact fading case.} The frozen Figure~3
    reference shows all 27 absolute-magnitude observations and error bars, the
    author-selected weighted fit, the impact-time marker, and the pre-impact
    magnitude with its uncertainty band. The astronomical magnitude axis is
    reversed, so upward movement denotes greater brightness.}
    \label{fig:case-dart}
\end{figure*}

\paragraph{Reference evidence chain.}
The executable analysis retains the author's weighting
(\texttt{sigma} set to squared magnitude error with
\texttt{absolute\_sigma=True}), fits the
21 selected observations, solves its crossing with the pre-impact magnitude,
and converts the aperture geometry and elapsed time to velocity. The fitted
observations and reference quantities are shown in Figure~\ref{fig:case-dart}.
The analysis recomputes a fading rate of $0.0579$ magnitude day$^{-1}$, a
crossing at 23.69 days, and a projected velocity of
$0.366\,\mathrm{m\,s^{-1}}$. The paper reports
$23.7\pm0.7$ days and $0.37\pm0.01\,\mathrm{m\,s^{-1}}$; the construction of
those two published uncertainties is not fully exposed by the plotting tables
and is therefore retained as source context rather than independently scored.

\FloatBarrier

\end{document}